\documentclass[10pt, a4paper]{article}

\usepackage[final]{lrec-coling2024} 

\usepackage{microtype}

\usepackage{times}
\usepackage{latexsym}
\usepackage{epsfig}
\usepackage{graphicx}

\usepackage{verbatim}

\usepackage{amsmath}
\usepackage{amssymb}
\usepackage{comment}
\usepackage{booktabs}
\usepackage{float}
\usepackage{multirow}
\usepackage{enumitem}
\usepackage{caption, subcaption, overpic, textpos}

\usepackage{xcolor}
\usepackage{colortbl}
\usepackage{makecell}
\definecolor{ours-highlight}{rgb}{0.86, 0.82, 1.0}
\definecolor{darkgray}{rgb}{0.66, 0.66, 0.66}
\definecolor{Gray}{gray}{0.9}

\title{Unifying Latent and Lexicon Representations for Effective Video-Text Retrieval}

\name{
\begin{tabular}{c}
Haowei Liu$^{1,2*}$, Yaya Shi$^{3*}$, Haiyang Xu$^{4\dagger}$, Chunfeng Yuan$^{1\dagger}$, Qinghao Ye$^{4}$\\
Chenliang Li$^{4}$, Ming Yan$^{4}$, Ji Zhang$^{4}$, Fei Huang$^{4}$, Bing Li$^{1}$, Weiming Hu$^{1,2,5}$\\ 
\end{tabular}
} 

\address{
$^{1}$MAIS, Institute of Automation, Chinese Academy of Sciences, China \\
$^{2}$School of Artificial Intelligence, University of Chinese Academy of Sciences, China \\
$^{3}$University of Science and Technology of China
\ $^{4}$Alibaba Group \\
$^{5}$School of Information Science and Technology, ShanghaiTech University, China \\
liuhaowei2019@ia.ac.cn, shiyaya@mail.ustc.edu.cn \\
\{shuofeng.xhy, ym119608\}@alibaba-inc.com, \{cfyuan, bli, wmhu\}@nlpr.ia.ac.cn \\
}

\abstract{
In video-text retrieval, most existing methods adopt the dual-encoder architecture for fast retrieval, which employs two individual encoders to extract global latent representations for videos and texts. However, they face challenges in capturing fine-grained semantic concepts. In this work, we propose the UNIFY framework, which learns lexicon representations to capture fine-grained semantics and combines the strengths of latent and lexicon representations for video-text retrieval. Specifically, we map videos and texts into a pre-defined lexicon space, where each dimension corresponds to a semantic concept. A two-stage semantics grounding approach is proposed to activate semantically relevant dimensions and suppress irrelevant dimensions. The learned lexicon representations can thus reflect fine-grained semantics of videos and texts. Furthermore, to leverage the complementarity between latent and lexicon representations, we propose a unified learning scheme to facilitate mutual learning via structure sharing and self-distillation. Experimental results show our UNIFY framework largely outperforms previous video-text retrieval methods, with 4.8\% and 8.2\% Recall@1 improvement on MSR-VTT and DiDeMo respectively. Code and pre-trained models will be publicly available at \href{https://github.com/auhowielau/UNIFY}{https://github.com/auhowielau/UNIFY}.
\\ \newline  }

\begin{document}
\maketitleabstract

\section{Introduction}

\let\thefootnote\relax\footnotetext{$^*$Equal contribution.

\ \ \ $^\dagger$Corresponding authors.}

Video-text retrieval is a crucial task with wide practical applications.
Recently, pre-training to learn transferable cross-modal representations has gradually become the paradigm of this field \cite{bain2021frozen,li2022alpro,bai2022lat,oa-trans,ge2022bridgeformer,Ge2022MILESVB}.
To achieve fast retrieval, most methods adopt the dual-encoder architecture.
It employs two individual encoders for video and text feature extraction respectively, and uses contrastive learning for cross-modal alignment.

As dual-encoder models compress a video (or text) into a latent vector, cross-modal interaction and alignment are solely based on such coarse-grained global representations.
Therefore, it's challenging for them to capture fine-grained semantic concepts such as objects and actions.
To tackle this, some methods (\textit{e.g.} \citealp{li2022alpro}) employ extra interaction modules to
enhance global latent representations.
However, it deprives the model's ability of fast retrieval.
BridgeFormer \cite{ge2022bridgeformer} discards the extra module after training, and thus the fine-grained interaction ability cannot be well transferred to the model when inference.

\begin{figure}
    \centering
    \includegraphics[width=1.0\linewidth]{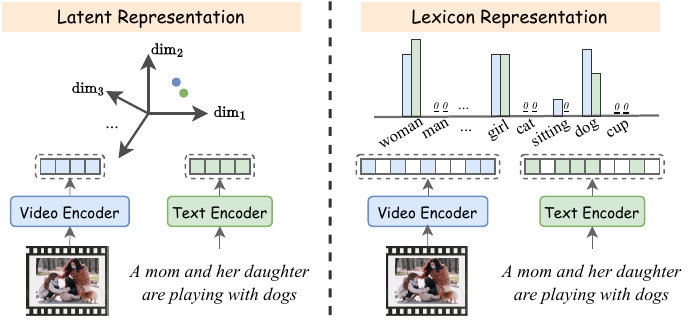}
\caption{Comparison of latent and lexicon representations.
The dimensions of latent representations have no explicit meanings.
In contrast, each dimension of lexicon representations corresponds to a semantic concept, where semantically relevant dimensions are activated (\textit{e.g.} woman and dog) while semantically irrelevant dimensions are suppressed (\textit{e.g.} cat and cup).}
\label{fig:intro}
\end{figure}

In this work, we present a novel \textbf{UNIFY} framework for unified video-text retrieval.
It learns lexicon representations of videos and texts to capture fine-grained semantics, and combines the strengths of latent and lexicon representations for effective cross-modal retrieval.
Firstly, we define a lexicon space where each dimension corresponds to a semantic concept represented by a word.
As shown in Figure \ref{fig:intro}, videos and texts are mapped into this space to obtain lexicon representations.
To capture fine-grained semantic information, we propose a two-stage semantics grounding approach to activate semantically relevant dimensions and suppress semantically irrelevant dimensions.
Secondly, as latent representations summarize videos and texts from a global perspective, and lexicon representations excel at capturing fine-grained semantics, combining them can further improve the model's performance.
Thus, to better leverage their complementarity, we propose a unified learning scheme which facilitates mutual learning between them via structure sharing and self-distillation.

Specifically, inspired by SPLADE \cite{Formal2021SPLADESL}, we can ground texts to semantically relevant dimensions by resorting to a pre-trained BERT \cite{devlin2018bert} model and its masked language modeling (MLM) head.
However, it's much more intractable for videos to achieve this due to the giant gap between raw pixels and the lexicon space.
To address this, we propose a two-stage semantics grounding approach.
As initially videos have random distributions in the lexicon space, in stage one, we freeze the text encoder to avoid textual lexicon representations being corrupted in cross-modal alignment.
We map local video and text features into the lexicon space using the MLM head, and aggregate them to obtain video-level and text-level lexicon representations.
Contrastive learning is then applied to pull paired samples closer and push unpaired ones away.
In stage two, we jointly train both the video and text encoders for further cross-modal alignment.
Apart from video-text contrastive learning, we employ the MLM task in this stage to preserve textual semantics.

To leverage the complementarity between latent and lexicon representations, we propose a unified learning scheme.
Firstly, from a structure sharing perspective, the two types of representations share a stem video (or text) encoder in shallow layers to promote knowledge sharing and transfer.
Meanwhile, representation-specific encoders are adopted in deep layers to focus on global and fine-grained semantic information respectively.
Secondly, as learning lexicon representations is relatively more challenging, we utilize latent representations to provide additional supervision information from a different perspective via self-distillation.
Specifically, we employ the similarity scores computed from latent representations as soft labels for the contrastive learning of lexicon representations.
Through the proposed unified learning scheme, latent and lexicon representations can benefit from each other, and are unified to form an effective video-text retriever.

Experimental results demonstrate the proposed lexicon representations can capture fine-grained semantics effectively.
Moreover, our UNIFY framework combines the strengths of latent and lexicon representations, and largely outperforms previous state-of-the-art methods in video-text retrieval.

Our contributions can be summarized as follows:
\begin{itemize}
    \item We present a novel UNIFY framework which unifies global latent representations and fine-grained lexicon representations for effective video-text retrieval.
    \item We propose a two-stage semantics grounding approach to ground videos and texts into semantically relevant dimensions, and a unified learning scheme to leverage the complementarity of latent and lexicon representations. 
    \item Experimental results show our model well captures fine-grained semantics and largely surpasses previous video-text retrieval methods. 
\end{itemize}

\section{Related Work}
\subsection{Video-Text Retrieval}
Recently, pre-training to learn transferable cross-modal representations has been popular in both image-text retrieval \cite{xu2021e2e,li2022mplug,xu2023mplug} and video-text retrieval \cite{miech2020milnce,bain2021frozen, xu2021videoclip,ge2022bridgeformer,Ge2022MILESVB}.
However, dual encoder methods have shortcomings in understanding the fine-grained alignment between video and text, which is crucial for accurate video-text retrieval. 
Currently, there are two ways to solve this problem. 
The first~\cite{li2022alpro, ge2022bridgeformer} involves using an additional fusion encoder to model fine-grained cross-modal interactions. 
However, the features cannot be pre-cached in this kind of structure and limit the model's fast retrieval ability.
The second one takes some learning strategies based on the dual encoder, such as MILES~\cite{Ge2022MILESVB} introduces masked image modeling task to inject the fine-grained semantics into global representation. Nevertheless, the fine-grained semantics are learned in an implicit way which may not be the optimal approach.
We believe that an explicit fine-grained representation will facilitate better retrieval performance. Therefore, in this paper, we introduce an efficient and effective method UNIFY by introducing a specific representation branch - lexicon one to capture fine-grained semantics, and we use a unifying scheme to promote collaboration between the fine-grained lexicon and original global latent representation.

\begin{figure*}
    \centering
    \includegraphics[width=0.98\linewidth]{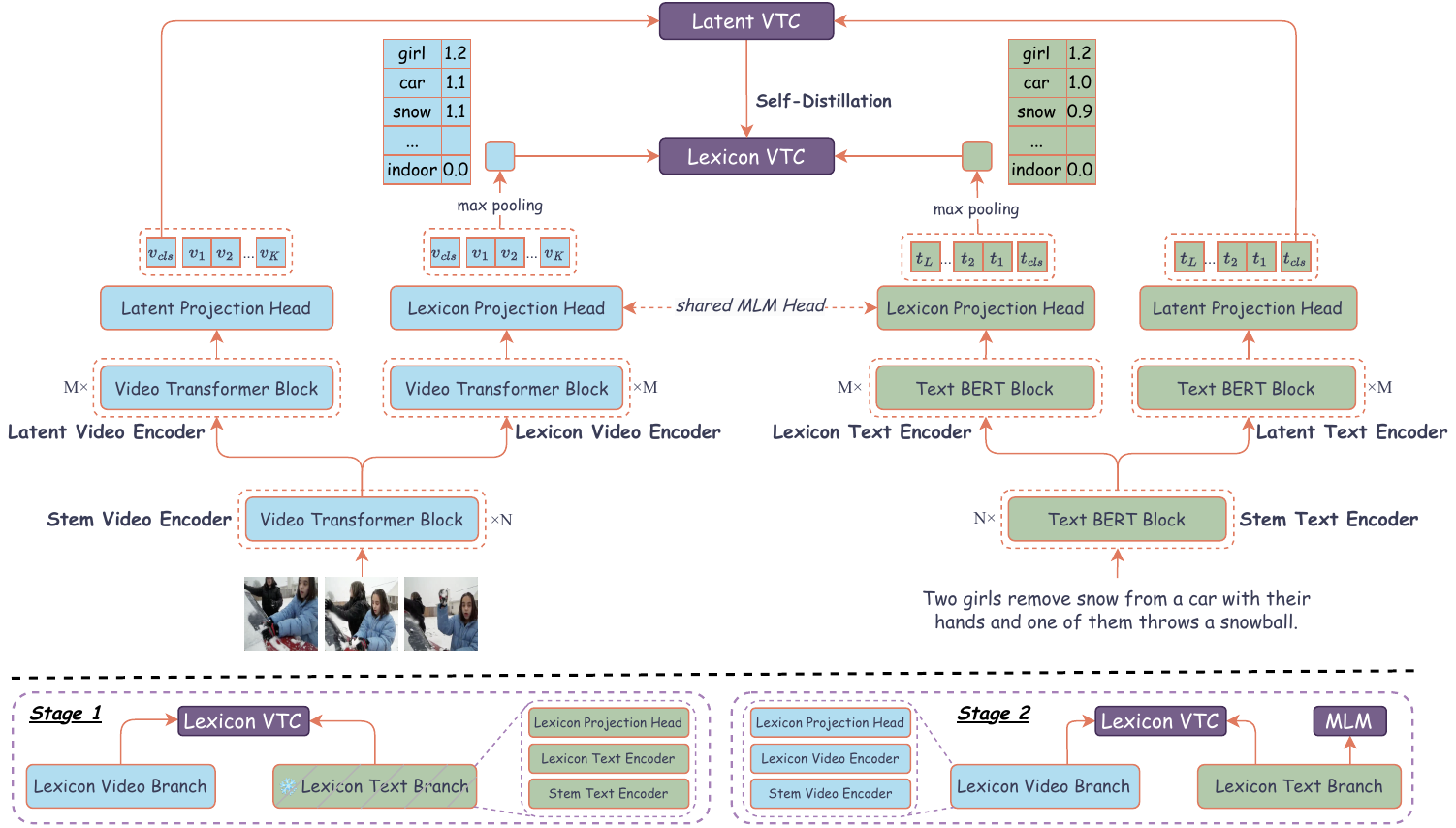}
\caption{
Overview of our proposed UNIFY framework.
The whole model consists of two streams for video and text respectively, each including a stem encoder, two representation-specific encoders and two projection heads.
For lexicon representation learning, we propose a two-stage semantics grounding approach (Section \ref{subsec:two_stage}).
Furthermore, we unify the latent and lexicon representations via structure sharing and self-distillation (Section \ref{subsec:unified_learning}). VTC stands for video-text contrastive learning.}
\label{fig:Main}
\end{figure*}

\subsection{Lexicon Representation}
The concept of lexicon representation was initially introduced by Vector Space Model \cite{salton1975vector}, which represents a
document as a vector in a vector space. Each dimension corresponds to a term in the vocabulary, whose value indicates the importance of the term in the document.
As pre-trained language models have gained popularity, neural network-based lexicon methods~\cite{Bai2020SparTermLT, Formal2021SPLADESL, Formal2021SPLADEVS, DBLP:conf/sigir/LassanceC22, Shen2022UnifieRAU} have achieved progress.
Our method is inspired by the neural-based lexicon representation methods. However, our task involves both video and text modalities.
Though text data can be naturally projected into the lexicon space by directly taking the pre-trained language model as the text encoder, it's much more intractable for videos to achieve this due to the giant modality gap between video and text.
To tackle this, in this paper, we propose a two-stage semantics grounding approach to learn lexicon representations of videos.

\section{Method}

\subsection{Overview}

\label{subsec:overview}
As Figure \ref{fig:Main} shows, we propose the UNIFY framework to unify latent and lexicon representations for effective video-text retrieval.
The architecture of UNIFY can be roughly divided into two parts, \textit{i.e.} the video stream and the text stream, to extract video and text representations respectively.
Both parts consist of five elements, \textit{i.e.}, a stem encoder $E_{stem}(\cdot)$ shared by the latent and lexicon representations, two representation-specific encoders $E_{lat}(\cdot)$, $E_{lex}(\cdot)$ and two corresponding projection heads $H_{lat}(\cdot)$, $H_{lex}(\cdot)$.
Note that videos and texts share the same lexicon projection head to map both modalities into the same lexicon space.

It takes two steps to extract latent and lexicon representations from videos and texts.
Taking an input video $V$ for example, as the left part of Figure \ref{fig:Main} shows, \textbf{1)} we first use the stem video encoder to extract video features, which are then fed into the latent video encoder and lexicon video encoder to obtain the corresponding raw features:
\begin{equation}
    r_{lat}^v = E_{lat}^v (E_{stem}^v(V)) \in \mathbb{R}^{d},
\end{equation}
\begin{equation}
    r_{lex}^v = E_{lex}^v (E_{stem}^v(V)) \in \mathbb{R}^{K \times d},
\end{equation}
where $K$ is the number of local video features, and $d$ is the dimension of raw features.
Note that the latent encoder only outputs the global [CLS] feature,
while the lexicon encoder outputs all local raw features.
\textbf{2)} We employ the latent and lexicon projection heads to map raw features into the corresponding representation spaces.
For latent representations, this step can be formulated as:
\begin{equation}
    f_{lat}^v = H_{lat}^v (r_{lat}^v) \in \mathbb{R}^{\hat{d}},
\end{equation}
where $\hat{d}$ is the dimension of the latent space.
As for the lexicon space, each dimension of it corresponds to a semantic concept represented by a word. Denoting the lexicon size as $|\mathbb{W}|$, the lexicon projection can be formulated as:
\begin{equation}
    p_{lex}^v = H_{lex}^v (r_{lex}^v) \in \mathbb{R}^{K \times |\mathbb{W}|}.
\end{equation}
After getting the lexicon representations of local patches, we perform the $\mathrm{ReLU}(\cdot)$ activation function to suppress negative values to zero, and aggregate them to get the video-level lexicon representation by pooling operation:
\begin{equation}
    f_{lex}^v =  \operatorname{Pool} (\operatorname{ReLU} (p_{lex}^v)) \in \mathbb{R}^{|\mathbb{W}|}.
\end{equation}

As the right part of Figure \ref{fig:Main} shows, following a similar process, we obtain the latent and lexicon representation of an input text $T$:
\begin{equation}
    f_{lat}^t = H_{lat}^t ( E_{lat}^t (E_{stem}^t(T)) ) \in \mathbb{R}^{\hat{d}},
\end{equation}
\begin{equation}
    p_{lex}^t = H_{lex}^t ( E_{lex}^t (E_{stem}^t(T)) ) \in \mathbb{R}^{L \times |\mathbb{W}|},
\end{equation}
\begin{equation}
    f_{lex}^t =  \operatorname{Pool} (\operatorname{ReLU} (p_{lex}^t)) \in \mathbb{R}^{|\mathbb{W}|},
\end{equation}
where $L$ is the number of tokens of $T$.

After obtaining the latent and lexicon representations of videos and text, when inference, we utilize both types of representations to calculate the dot product similarity scores ($S_{lat}$ and $S_{lex}$) between video-text pairs.
Finally, we combine the two scores at a 1:1 ratio to serve as the final similarity score for ranking:
\begin{equation}
    S = S_{lat} + S_{lex}.
\end{equation}

\subsection{Two-stage Semantics Grounding}
\label{subsec:two_stage}
In order to capture fine-grained semantics, lexicon representations are required to satisfy two semantic constraints: \textbf{1)} dimensions semantically relevant to the video (or text) are activated with high values, and \textbf{2)} semantically irrelevant dimensions are suppressed to zero.
This makes the learning of lexicon representations quite challenging.
Luckily, for texts, we can resort to pre-trained language models (PLM) to achieve this goal.
PLMs such as BERT \cite{devlin2018bert} are trained with the masked language modeling (MLM) task, which can project the masked tokens to semantically relevant words.
Therefore, by reusing PLM and its MLM head, we can transform texts into lexicon representations that satisfies the semantic constraints.
However, it's much more intractable for videos to achieve the same goal, as video's raw pixels are significantly different from discrete words in both modality and semantics.
To tackle this, we propose a two-stage semantics grounding approach.

\vspace{0.4em}
\noindent \textbf{Stage 1.}
As initially videos are randomly distributed in the lexicon space, updating video and text encoders simultaneously may damage the lexicon distributions of texts.
Therefore, as the lower left part of Figure \ref{fig:Main} shows, we freeze the text encoders and ground videos into the lexicon space in stage 1.
We resort to the paired texts to acquire the information which dimensions of the lexicon space are relevant to the videos.
Specifically, we learn cross-modal semantic alignment in the lexicon space by optimizing a video-text contrastive (VTC) learning objective with Noise-Contrastive Estimation (NCE):
\begin{equation}
\begin{gathered}
\mathcal{L}_{\mathrm {VTC}}^{lex} =\sum_{i=1}^B \mathrm{NCE}\left({v}_i, {t}_i\right)+\sum_{i=1}^B \mathrm{NCE}\left({t}_i, {v}_i\right), \\
\mathrm{NCE}\left(x_i, y_i\right)=-\log \frac{\exp \left(x_i^T y_i / \tau\right)}{\sum_{j=1}^B \exp \left(x_i^T y_j / \tau\right)},
\end{gathered}
\label{equ:vtc_lex}
\end{equation}
where $v_i$ and $t_i$ are normalized lexicon representations of the $i$-th video and text in a batch. 
$B$ is batch size and $\tau$ is a temperature hyper-parameter.

\vspace{0.4em}
\noindent \textbf{Stage 2.}
As the lower right part of Figure \ref{fig:Main} shows, in stage 2, we jointly train the video and text encoders for bidirectional cross-modal alignment.
However, simply unfreezing the text encoders will deprive the semantic constraints and cause texts to drift in the lexicon space.
Therefore, apart from video-text contrastive learning, we adopt the masked language modeling (MLM) task in stage 2 
for preserving
textual semantics.
MLM recovers the masked tokens by reasoning contextual text, and thus encourages the model to project text tokens to lexicon dimensions corresponding to their semantics.
Denoting the text with tokens masked as $\hat{T}$ and the prediction probability of the masked tokens as $\boldsymbol{p}^{\mathrm {mask}}( \hat{T})$,
MLM loss can be formulated as:
\begin{equation}
\mathcal{L}_{\mathrm{MLM}}=\mathbb{E}_{\hat{T} \sim D}\left[\operatorname{CE}\left(\boldsymbol{y}^\mathrm{mask}, \boldsymbol{p}^{\mathrm {mask }}( \hat{T})\right)\right],
\end{equation}
where $\boldsymbol{y}^{\mathrm{mask}}$ is the ground truth one-hot vectors of the masked tokens. $D$ is the training dataset. $\mathrm{CE}$ stands for cross-entropy loss.

As the lexicon space is high-dimensional (\textit{e.g.} 30522-d), to avoid semantically relevant dimensions being overridden by massive non-zero values on those semantically irrelevant dimensions,
we introduce a FLOPs loss \cite{Paria2020FLOPs} to encourage the sparsity of the lexicon representations:
\begin{equation}
    \mathcal{L}_{\mathrm{FLOPs}} = \
 \sum_{k}^{|\mathbb{W}|} (\frac{1}{B} \sum_{i=1}^B {v}_i^{k} )^2 +
 \sum_{k}^{|\mathbb{W}|} (\frac{1}{B} \sum_{i=1}^B {t}_i^{k} )^2,
\end{equation}
where $v_i^k$ and $t_i^k$ are the activation values of the $k$-th dimension of the lexicon space.

The training objectives for learning lexicon representations in stage 1 and 2 are as follows:
\begin{equation}
    \mathcal{L}^{lex} = \begin{cases} \mathcal{L}_{\mathrm{VTC}}^{lex} + \beta \cdot \mathcal{L}_{\mathrm{FLOPs}}, & \mathrm {stage 1} \\ 
    \mathcal{L}_{\mathrm{VTC}}^{lex} + \beta \cdot \mathcal{L}_{\mathrm{FLOPs}} + \mathcal{L}_{\mathrm{MLM}}, & \mathrm {stage 2} \end{cases}
\end{equation}
where $\beta$ is the weight of the FLOPs loss.

\subsection{Unified Learning of Latent and Lexicon Representations}
\label{subsec:unified_learning}
Latent representations focus more on global content, while lexicon representations excel at capturing fine-grained semantics.
To leverage the complementarity of latent and lexicon representations, we propose a unified learning scheme to facilitate their mutual learning.

\vspace{0.4em}
\noindent \textbf{Structure sharing.}
Firstly, we unify the learning of latent and lexicon representations from a structure sharing perspective.
To combine the strengths of both representations, an intuitive way is to train two individual dual-encoder models for latent and lexicon representations respectively, and apply score-level fusion for retrieval.
However, this parallel architecture has two drawbacks. \textbf{1)} The number of parameters and computation cost are doubled.
\textbf{2)} Lacking interaction prevents them from mutual learning, and thus can't achieve the optimal performance of unified retrieval.

As shown in Figure \ref{fig:Main}, we instead propose a unified architecture, where the latent and lexicon branches share the same stem video (or text) encoder.
As the two types of representations focus on information of different granularities, sharing shallow layers promotes knowledge sharing and transfer between them during the learning process.
On the other hand, if the whole video (or text) encoder is shared, the raw features input to the latent and lexicon projection heads will be identical, which inevitably harms the complementarity between the two representation types.
Therefore, in each stream, we introduce two representation-specific encoders on top of the stem encoder.
The latent and lexicon encoders are optimized by different training objectives, and thus can learn visual and textual information of different granularities.

\vspace{0.4em}
\noindent \textbf{Self-distillation.}
Secondly, we use self-distillation to facilitate knowledge transfer from latent to lexicon representations. 
On the one hand, learning lexicon representations are more challenging due to the semantic constraints.
On the other hand, while freezing the lexicon text branch in stage 1 (Section \ref{subsec:two_stage}) avoids the textual lexicon distributions being corrupted, it also to some extent limits the ability of lexicon representations.
As they have different focuses, the knowledge from latent representations can provide extra supervision for lexicon representation learning. 
Specifically, for each type of representations, we obtain the similarity scores between videos and texts by computing the dot product of their normalized representations.
Denote video-to-text and text-to-video similarity scores as $S_{v2t}$ and $S_{t2v}$.
We use the similarity scores of latent representations as soft labels to perform self-distillation on lexicon representations, and optimize a KL divergence ($D_{\mathrm{KL}}$) loss as follows:
\begin{equation}
    \mathcal{L}_{\mathrm{D}} = D_{\mathrm{KL}}(S_{v2t}^{lex} || S_{v2t}^{lat}) + D_{\mathrm{KL}}(S_{t2v}^{lex} || S_{t2v}^{lat}).
\end{equation}

Combining the self-distillation loss and the video-text contrastive loss of latent representations, the overall training objective of our UNIFY framework can be formulated as:
\begin{equation}
    \mathcal{L} = \mathcal{L}^{lex} + \mathcal{L}_{\mathrm{VTC}}^{lat} + \lambda \cdot \mathcal{L}_{\mathrm{D}},
\end{equation}
where $\mathcal{L}_{\mathrm{VTC}}^{lat}$ has the same form as $\mathcal{L}_{\mathrm{VTC}}^{lex}$ in Equation \ref{equ:vtc_lex}, and $\lambda$ is the weight of the self-distillation loss.
In practice, we linearly decrease $\lambda$ during training, which avoids harming the complementarity between latent and lexicon representations while facilitating lexicon representation learning.

\begin{table*}[t]
\centering
\scalebox{0.78}{
\begin{tabular}{lcccc|cccc}

    \toprule[1pt]
    \multirow{2}{*}{Method} & \multirow{2}{*}{Year} & \multirow{2}{*}{Video Input} & \multirow{2}{*}{Pre-train Dataset} & \multirow{2}{*}{Pairs} & \multicolumn{4}{c}{\textbf{MSR-VTT}} \\ 
    \cmidrule(l){6-9} 
     & & & & & R@1$\uparrow$ & R@5$\uparrow$ & R@10$\uparrow$ & MedR$\downarrow$ \\
    
    \midrule
    
    \multicolumn{9}{c}{Zero-Shot} \\
    \midrule
    \multirow{1}*{SupportSet~\cite{support}}&2021&R(2+1)D-34&HowTo100M&120M&12.7&27.5&36.2&24.0\\
    \multirow{1}*{Frozen~\cite{bain2021frozen}}&2021&Raw Videos&CC3M, WebVid-2M&5.5M&18.7&39.5&51.6&10.0\\
    \multirow{1}*{AVLnet~\cite{avlnet}}&2021&ResNeXt-101&HowTo100M&120M&19.6&40.8&50.7&9.0\\
    \multirow{1}*{RegionLearner~\cite{yan2023video}}&2023&Raw Videos&CC3M, WebVid-2M&5.5M& 22.2 & 43.3 & 52.9 & 8.0 \\
    \multirow{1}*{LaT~\cite{bai2022lat}}&2022&Raw Videos&CC3M, WebVid-2M&5.5M&23.4&44.1&53.3&8.0\\
    \multirow{1}*{MILES~\cite{Ge2022MILESVB}}&2022&Raw Videos&CC3M, WebVid-2M&5.5M&26.1&47.2&56.9&7.0\\
    \multirow{1}*{BridgeFormer~\cite{ge2022bridgeformer}}&2022&Raw Videos&CC3M, WebVid-2M&5.5M&26.0&46.4&56.4&7.0\\
    \multirow{1}*{TCP~\cite{zhang2023exploring}}&2023&Raw Videos&CC3M,  WebVid-2M&5.5M&  26.8 & 48.3  & 57.6  & 7.0  \\
    \midrule
    \multirow{1}*{UNIFY-Latent}&2023&Raw Videos&CC3M, WebVid-2M&5.5M & 28.4 & 49.4 & 59.4 & 6.0 \\
    \multirow{1}*{UNIFY-Lexicon}&2023&Raw Videos&CC3M, WebVid-2M&5.5M & 28.0 & 50.2 & 59.8 & 5.0 \\
    \rowcolor{ours-highlight}
    \multirow{1}*{\textbf{UNIFY}}&2023&Raw Videos&CC3M, WebVid-2M&5.5M&\textbf{29.0}&\textbf{51.7}&\textbf{60.1}&\textbf{5.0} \\		
    \midrule
    \multicolumn{9}{c}{Fine-Tuning} \\
    \midrule
    \multirow{1}*{SupportSet~\cite{support}}&2021&R(2+1)D-34&HowTo100M&120M&30.1&58.5&69.3&3.0\\	
    \multirow{1}*{VideoCLIP~\cite{xu2021videoclip}}&2021&S3D&HowTo100M&110M&30.9&55.4&66.8&-\\
    \multirow{1}*{Frozen~\cite{bain2021frozen}}&2021&Raw Videos&CC3M, WebVid-2M&5.5M&31.0&59.5&70.5&3.0\\
    \multirow{1}*{ALPRO~\cite{li2022alpro}}&2022&Raw Videos&CC3M, WebVid-2M&5.5M&33.9&60.7&73.2&3.0\\
    \multirow{1}*{LaT~\cite{bai2022lat}}&2022&Raw Videos&CC3M, WebVid-2M&5.5M&35.3&61.3&72.9&3.0\\
    \multirow{1}*{OA-Trans~\cite{oa-trans}}&2022&Raw Videos&CC3M, WebVid-2M&5.5M&35.8&63.4&76.5&3.0\\
    \multirow{1}*{RegionLearner~\cite{yan2023video}}&2023&Raw Videos&CC3M, WebVid-2M&5.5M& 36.3 & 63.9 & 72.5 & 3.0 \\
    \multirow{1}*{MILES~\cite{Ge2022MILESVB}}&2022&Raw Videos&CC3M,  WebVid-2M&5.5M&37.7&63.6&73.8&3.0\\		
    \multirow{1}*{BridgeFormer~\cite{ge2022bridgeformer}}&2022&Raw Videos&CC3M,  WebVid-2M&5.5M&37.6&64.8&75.1&3.0\\
    \multirow{1}*{TCP~\cite{zhang2023exploring}}&2023&Raw Videos&CC3M,  WebVid-2M&5.5M& 38.0 & 65.5 & 76.4 & 3.0 \\
    \midrule
    \multirow{1}*{UNIFY-Latent}&2023&Raw Videos&CC3M, WebVid-2M&5.5M & 40.1 & 66.3 & 75.0 & 2.0\\
    \multirow{1}*{UNIFY-Lexicon}&2023&Raw Videos&CC3M, WebVid-2M&5.5M & 40.8 & 68.9 & 78.2 & 2.0\\
    \rowcolor{ours-highlight}
    \multirow{1}*{\textbf{UNIFY}}&2023&Raw Videos&CC3M, WebVid-2M&5.5M&\textbf{42.8}&\textbf{68.8}&\textbf{78.8}&\textbf{2.0}\\			
    \bottomrule[1pt]
\end{tabular}}
\caption{Text-to-video retrieval results on MSR-VTT test set. ``Video Input'' lists the input to the video encoder, where ``Raw Videos'' means training on raw video frames without pre-extracted features.}
\label{tab:msrvtt}
\end{table*}

\section{Experiments}
\subsection{Experimental Setup}
\noindent \textbf{Pre-training Datasets.}
For a fair comparison, we follow Frozen \cite{bain2021frozen} and MILES \cite{Ge2022MILESVB} to adopt two pre-training datasets - Google Conceptual Captions (CC3M) \cite{sharma2018conceptual} with 3M image-text pairs, and WebVid-2M \cite{bain2021frozen} with 2.5M video-text pairs.

\vspace{0.4em}
\noindent \textbf{Downstream Tasks.}
\textbf{1) Text-to-video retrieval.}
We evaluate the text-to-video retrieval performance of our UNIFY model on four mainstream datasets, \textit{i.e.}, \text{MSR-VTT}~\cite{xu2016msrvtt}, \text{DiDeMo}~\cite{anne2017didemo}, \text{LSMDC}~\cite{rohrbach2015lsmdc} and \text{MSVD}~\cite{chen2011msvd}.
The evaluation adopts both zero-shot and fine-tuning setups, and uses Recall and Median Rank as metrics.
\textbf{2) Action Recognition.}
We also evaluate our model's performance in zero-shot action recognition, which can be regarded as video-to-text retrieval by describing videos with corresponding action classes following \cite{radford2021clip}.
Evaluation is conducted on the \text{HMDB51}~\cite{Kuehne2011hmdb51} and \text{UCF101}~\cite{Soomro2012ucf101} datasets.
Both datasets are divided into three training/test splits.

\vspace{0.4em}
\noindent \textbf{Implementation Detail.}
We instantiate the video encoder with the Timesformer~\cite{bertasius2021space} model, and initialize the parameters of spatial attention blocks by reusing ViT-B/16 weights pre-trained on ImageNet-21K~\cite{ridnik2021imagenet}.
As for the text encoder, we use the BERT$_{base}$ ~\cite{devlin2018bert} model for initialization, and take its word embedding vocabulary as our lexicon.
In default, the number N of shared stem blocks is set as 9, and M of representation-specific blocks is set as 3.
We use 8 NVIDIA A100 GPUs for pre-training and 8 NVIDIA V100 GPUs for fine-tuning.
We pre-trained UNIFY for a total of 10 epochs, during which the text encoder was frozen for the first 3 epochs. 
We utilize the AdamW \cite{Loshchilov2019admw} optimizer with a weight decay of 0.05 and batch size of 512. The learning rate was initially raised to 1e-4 in the first epoch and then decayed based on a cosine schedule. 
We randomly select 4 frames per video and resize to 224x224.
We set the mask ratio of MLM task as 15\%, and empirically set the weight of the FLOPs loss as 1e-4.
As for the weight $\lambda$ of self-distillation loss, we linearly decrease it from 1 to 0.

\begin{table*}[]
\centering
	\scalebox{0.68}{
	\begin{tabular}{lcccc|cccc|cccc}

        \toprule[1pt]
 	\multirow{2}{*}{Method}& \multicolumn{4}{c|}{\textbf{DiDeMo}} & \multicolumn{4}{c|}{\textbf{LSMDC}} & \multicolumn{4}{c}{\textbf{MSVD}} \\
	\cmidrule(l){2-13} 
	 & R@1$\uparrow$ & R@5$\uparrow$ & R@10$\uparrow$ & MedR$\downarrow$  & R@1$\uparrow$ & R@5$\uparrow$ & R@10$\uparrow$ & MedR$\downarrow$  & R@1$\uparrow$ & R@5$\uparrow$ & R@10$\uparrow$ & MedR$\downarrow$ \\
 
        \midrule
        \multicolumn{13}{c}{Zero-Shot} \\
        \midrule
        \multirow{1}*{Frozen~\cite{bain2021frozen}}&21.1&46.0&56.2&7.0&9.3&22.0&30.1&51.0&33.7&64.7&76.3&3.0\\
            \multirow{1}*{LaT~\cite{bai2022lat}}&22.6&45.9&58.9&7.0&-&-&-&-&36.9&68.6&81.0&2.0 \\
        \multirow{1}*{OA-Trans~\cite{oa-trans}}&23.5&50.4&59.8&6.0&-&-&-&-&39.1&68.4&80.3&2.0 \\			
            \multirow{1}*{MILES~\cite{Ge2022MILESVB}}&27.2&50.3&63.6&5.0&11.1&24.7&30.6&50.7&44.4&76.2&87.0&2.0\\
        \multirow{1}*{BridgeFormer~\cite{ge2022bridgeformer}}&25.6&50.6&61.1&5.0&12.2&25.9&32.2&42.0&43.6&74.9&84.9&2.0\\
        \midrule
        \multirow{1}*{UNIFY-Latent} & 27.3 & 53.8 & 63.5 & 4.0 & 12.4 & 28.4 & 36.1 & 30.5 & 46.4 & 77.0 & 87.0 & 2.0   \\
        \multirow{1}*{UNIFY-Lexicon} & 28.3 & 53.4 & 63.8 & 4.0 & 12.6 & 27.2 & 36.6 & 28.0 & 48.3 & 77.2 & 86.7 & 2.0  \\
        \rowcolor{ours-highlight}
        \multirow{1}*{\textbf{UNIFY}}&\textbf{29.6}&\textbf{55.5}&\textbf{66.0}&\textbf{4.0}&\textbf{14.1}&\textbf{30.4}&\textbf{37.5}&\textbf{25.0}&\textbf{48.1}&\textbf{79.7}&\textbf{87.2}&\textbf{2.0}\\
        \midrule
        \multicolumn{13}{c}{Fine-Tuning} \\
        \midrule 
            \multirow{1}*{CipBert~\cite{lei2021clipbert}}&20.4&48.0&60.8&6.0&-&-&-&-&-&-&-&- \\
        \multirow{1}*{RegionLearner~\cite{yan2023video}}& 32.5 & 60.8 & 72.3 & 3.0 & 17.1 & 32.5 & 41.5 & 18.0 & 44.0 & 74.9 & 84.3 & 2.0 \\
            \multirow{1}*{LaT~\cite{bai2022lat}}&32.6&61.3&71.6&3.0&-&-&-&-&40.0&74.6&84.2&2.0 \\
            \multirow{1}*{OA-Trans~\cite{oa-trans}}&34.8&64.4&75.1&3.0&18.2&34.3&43.7&18.5&-&-&-&- \\		
            \multirow{1}*{ALPRO~\cite{li2022alpro}}&35.9&67.5&78.8&3.0&-&-&-&-&-&-&-&- \\
        \multirow{1}*{MILES~\cite{Ge2022MILESVB}}&36.6&63.9&74.0&3.0&17.8&35.6&44.1&15.5&53.9&83.5&90.2&1.0\\		    
        \multirow{1}*{BridgeFormer~\cite{ge2022bridgeformer}}&37.0&62.2&73.9&3.0&17.9&35.4&44.5&15.0&52.0&82.8&90.0&1.0\\
        \midrule
        \rowcolor{Gray}
        \multirow{1}*{CLIP4Clip~\cite{luo2022clip4clip}}&43.4&70.2&80.6&2.0&22.6&41.0&49.1&11.0&46.2&76.1&84.6&2.0\\	
        \rowcolor{Gray}
        \multirow{1}*{CenterCLIP~\cite{zhao2022centerclip}}&-&-&-&-&21.9&41.1&50.7&10.0&47.6&77.5&86.0&2.0\\	
        \rowcolor{Gray}
        \multirow{1}*{DiCoSA~\cite{jin2023text}}&45.7&74.6&83.5&2.0&25.4&43.6&54.0&8.0&47.4&76.8&86.0&2.0\\	
        \midrule
        \multirow{1}*{UNIFY-Latent} & 40.7 & 68.5 & 80.0 & 2.0 & 22.8 & 42.2 & 50.8 & 10.0 & 55.5 & 85.2 & 91.6 & 1.0  \\
        \multirow{1}*{UNIFY-Lexicon} & 44.6 & 73.5 & 82.8 & 2.0 & 24.5 & 44.6 & 54.4 & 8.0 & 55.1 & 86.7 & 93.0 & 1.0 \\
        \rowcolor{ours-highlight}
        \multirow{1}*{\textbf{UNIFY}}&\textbf{45.2}&\textbf{74.0}&\textbf{83.2}&\textbf{2.0}&\textbf{24.5}&\textbf{46.3}&\textbf{55.0}&\textbf{7.0}&\textbf{57.7}&\textbf{86.8}&\textbf{92.9}&\textbf{1.0} \\			
        \bottomrule[1pt]
	\end{tabular}
        }
 	\caption{Experimental results of text-to-video retrieval on the DiDeMo, LSMDC and MSVD datasets.} 
	\label{tab:multi-datasets}
\end{table*}

\begin{table}\centering
\scalebox{0.52}{
\begin{tabular}{lcccc|cccc}

\toprule[1pt]
\multirow{2}{*}{Method}&\multicolumn{4}{c|}{\textbf{HMDB51}}&\multicolumn{4}{c}{\textbf{UCF101}} \\
\cmidrule(l){2-9} 
 &S1&S2&S3&Mean&S1&S2&S3&Mean\\

\midrule 
ClipBert~\cite{lei2021clipbert}&20.0&22.0&22.3&21.4&27.5&27.0&28.8&27.8\\
Frozen~\cite{bain2021frozen}&27.5&28.3&27.7&27.8&45.4&44.7&47.7&45.9\\
MILES~\cite{Ge2022MILESVB}&{38.4}&{38.6}&{37.8}&{38.3}&{51.8}&{53.4}&{52.8}&{52.7}\\	
BridgeFormer~\cite{ge2022bridgeformer}&{38.0}&{36.1}&{39.1}&{37.7}&{51.1}&{54.3}&{ \textbf{53.8} }&{53.1}\\	
\rowcolor{ours-highlight}
\textbf{UNIFY} (Ours) &\textbf{38.9}&\textbf{39.6}&\textbf{39.9}&\textbf{39.5}&\textbf{53.3}&\textbf{55.0}& 53.0 &\textbf{53.8}\\	
\bottomrule[1pt]
\end{tabular}}
\caption{Top-1 accuracy of zero-shot action recognition. ``S'' denotes different testing splits, and ``Mean'' is the averaged result over three splits.} 
\label{tab:zero_action}
\end{table}

\subsection{Main Results}
\noindent\textbf{Evaluation on Video-Text Retrieval.}
Table \ref{tab:msrvtt} shows the performance on the MSR-VTT dataset under zero-shot and fine-tuning settings. UNIFY-Latent and UNIFY-Lexicon are the retrieval results of the two representation types in our model, and UNIFY denotes the score-level fusion results of them. Based on the retrieval results, we make the following observations:
1) UNIFY-Lexicon surpasses UNIFY-Latent in almost all metrics, validating that our proposed two-stage semantics grounding approach can capture fine-grained semantic concepts and boost retrieval performance effectively.
2) UNIFY-Latent also significantly outperforms existing latent-representation-based methods, showing the unified learning scheme allows latent and lexicon representations to benefit from each other and improves performance.
3) The combination of latent and lexicon representations largely improves performance, demonstrating our UNIFY can combine the strengths of both representation types effectively.
Overall, under the fine-tuning setting, UNIFY significantly outperforms TCP \cite{zhang2023exploring} by 4.8\%.

Table \ref{tab:multi-datasets} shows the retrieval performance on the DiDeMo, LSMDC and MSVD datasets. Similar to MSR-VTT, our UNIFY model surpasses previous methods in almost all metrics. Notably, under the fine-tuning setting, we outperform BridgeFormer \cite{ge2022bridgeformer} by 8.2\% on the challenging DiDeMo dataset, which consists of longer videos and more complex semantic concepts compared to other datasets. The result validates the effectiveness of our approach.
Another point worth noting is that on these three datasets, we list the performance of CLIP-based methods such as CLIP4Clip \cite{luo2022clip4clip}, CenterCLIP \cite{zhao2022centerclip} and DiCoSA \cite{jin2023text}. These models are initialized using the parameters of the CLIP \cite{radford2021clip} model, which is pre-trained on 400M image-text pair data (70$\times$ more than our pre-training data). Our method even surpasses them or achieves comparable performance, which further demonstrates the superiority of our method.

\vspace{0.4em}
\noindent \textbf{Evaluation on Action Recognition.}
Table \ref{tab:zero_action} presents the zero-shot action recognition performance on HMDB51 and UCF101, which can be regarded as video-to-text retrieval. Except for UCF101 split 3, we outperform previous methods in all splits. This verifies that our UNIFY model can learn cross-modal representations that generalize well to the task of action recognition.

\begin{figure}[t]
\centering
\includegraphics[width=\linewidth]{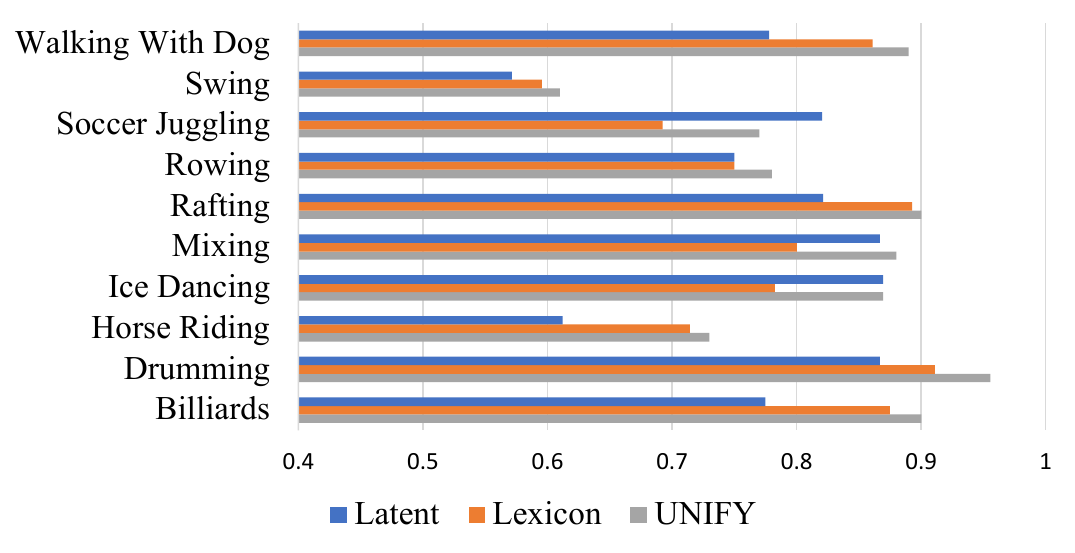}
\caption{Zero-shot results of latent and lexicon representations on UCF101.}
\label{fig:action}
\end{figure}

\begin{figure}[t]
\centering
\includegraphics[width=0.95\linewidth]{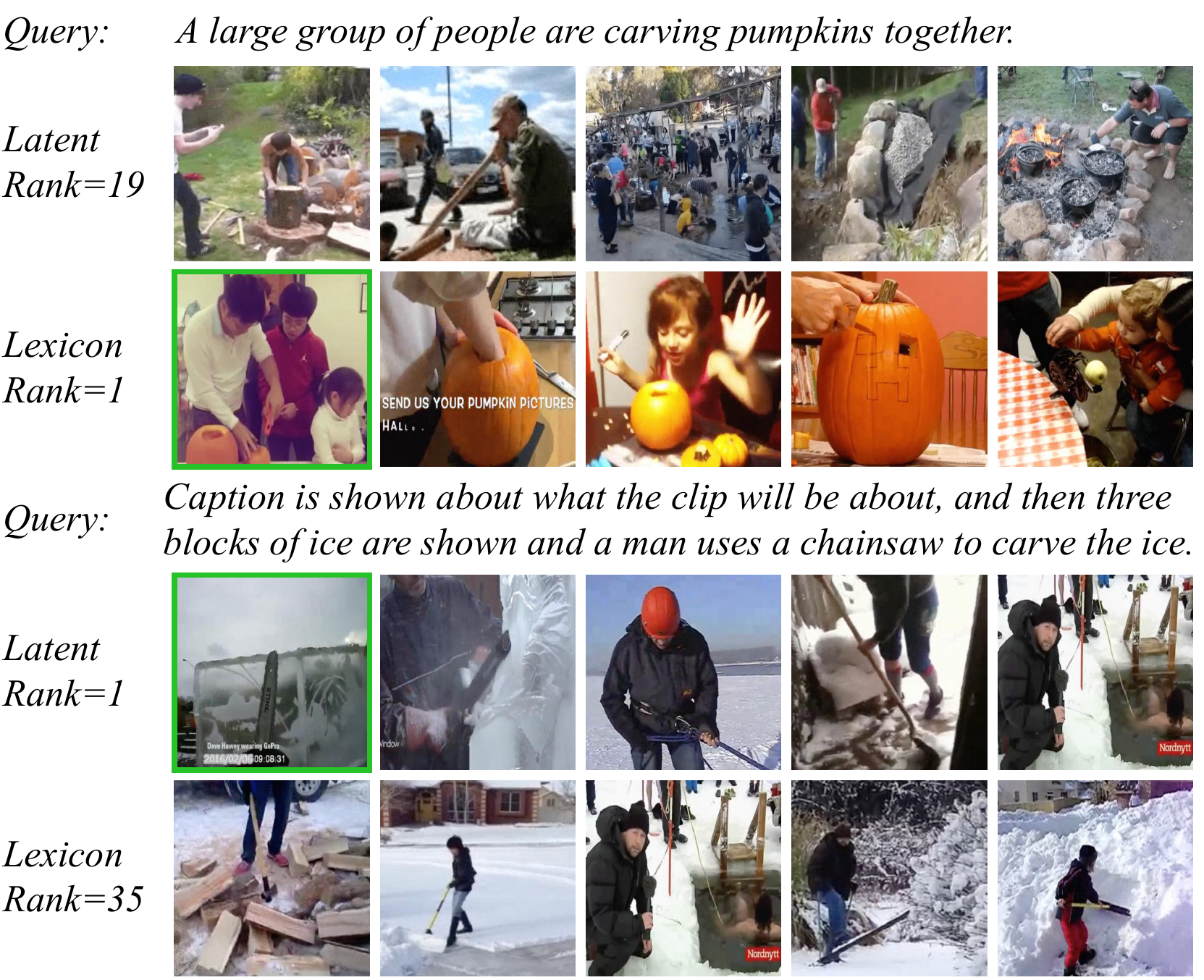}
\caption{Retrieval results of two queries using latent and lexicon representations. Each row presents the top-5 ranked videos.}
\label{fig:case_study}
\end{figure}

\begin{figure*}
    \centering
    \includegraphics[width=0.99\linewidth]{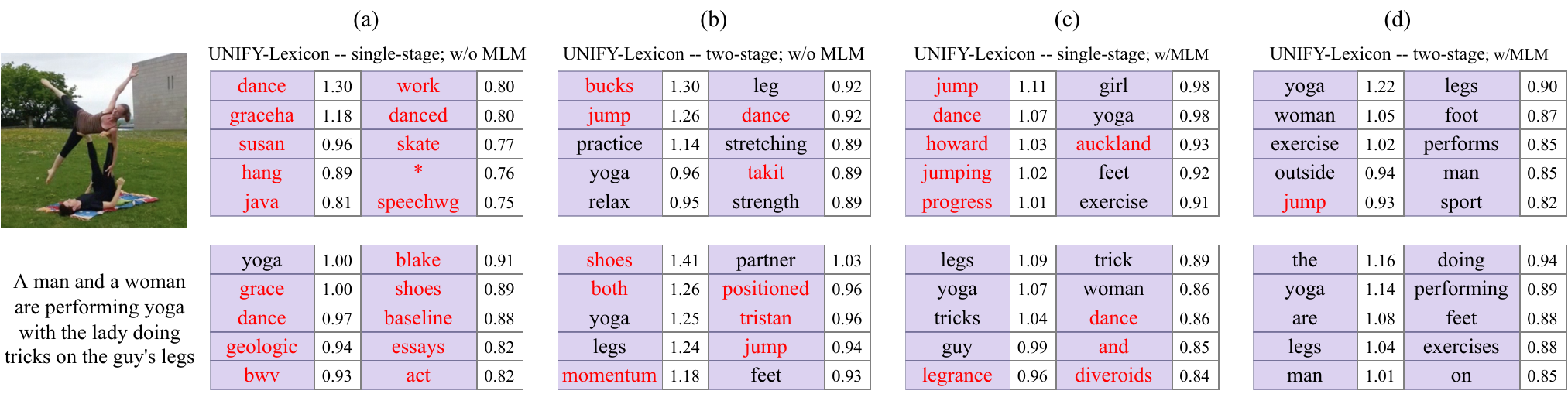}
\caption{Top-10 activated lexicon dimensions of four variants of UNIFY-Lexicon. Words that are semantically irrelevant to the video (or text) are highlighted in red color.}
\label{fig:Two_stage_semantics_grounding_ablation}
\end{figure*}

\begin{table*}[]
\centering
\scalebox{0.62}{%
\begin{tabular}{c|cc|cccc|cccc|cccc}
\toprule
\multirow{2}{*}{\#Line} & \multirow{2}{*}{sharing stem encoder} &
  \multirow{2}{*}{self-distillation} &
  \multicolumn{4}{c|}{UNIFY-Latent} &
  \multicolumn{4}{c|}{UNIFY-Lexicon} &
  \multicolumn{4}{c}{$\text{UNIFY}$} \\ 
  \cmidrule{4-15} 
       &      &     & R@1$\uparrow$ & R@5$\uparrow$ & R@10$\uparrow$ & MedR$\downarrow$ & R@1$\uparrow$ & R@5$\uparrow$ & R@10$\uparrow$ & MedR$\downarrow$ & R@1$\uparrow$ & R@5$\uparrow$ & R@10$\uparrow$ & MedR$\downarrow$ \\ 
\midrule
A & $\times$ & $\times$ & 24.7 & 47.8 & 56.9 & 6.0 & 26.6 & 48.1 & 57.2 & 6.0 & 26.8 & 48.3 & 57.9 & 6.0 \\
B & $\checkmark$ & $\times$ & 26.9 & 48.2 & 58.7 & 6.0 & 27.5 & 48.5 & 58.9 & 6.0 & 28.2 & 49.5 & 59.3 & 6.0 \\
\rowcolor{ours-highlight}C & $\checkmark$ & $\checkmark$ & \textbf{28.4} & \textbf{49.4} & \textbf{59.4} & \textbf{6.0} & \textbf{28.0} & \textbf{50.2} & \textbf{59.8} & \textbf{5.0} & \textbf{29.0} & \textbf{51.7} & \textbf{60.1} & \textbf{5.0} \\ 
\bottomrule

\end{tabular}%
}
\caption{Ablation study on the unified learning scheme which includes structure sharing and self-distillation. Zero-shot text-to-video retrieval results on MSR-VTT are reported.}
\label{tab:unify_ablation}
\end{table*}

\subsection{Complementarity between Latent and Lexicon Representations}

Figure \ref{fig:action} shows the zero-shot performance of several actions in UCF101. 
Lexicon representations grasp fine-grained object information more effectively, showing better performance on classes like ``Billiards" and ``Horse Riding".
On the other hand, latent representations beat lexicon representations on classes like ``Mixing" and ``Ice Dancing", which rely more on long-range action analysis.
Moreover, combining both types of representations leads to further improvement on most classes.

Figure \ref{fig:case_study} shows the retrieval results of two queries using latent and lexicon representations respectively, where "Rank=.." indicates the ranking of the ground truth video among all candidate videos.
In the first example, the lexicon representations capture the crucial detailed semantics of ``pumpkin", and thus successfully retrieves the correct video. The second query is longer and corresponds to a video with complex content. Though lexicon representations manage to capture some relevant semantics (\textit{e.g.} man and snow), aggregating local semantics fails to grasp the overall semantics of complex texts and videos. In contrast, latent representations summarize texts and videos from a global perspective and successfully retrieves the correct video.

\begin{table}[t]
\centering
\scalebox{0.62}{
\begin{tabular}{c|cc|cccc}
\toprule
\#Line & freezing strategy & MLM task  & R@1$\uparrow$ & R@5$\uparrow$ & R@10$\uparrow$ & MedR$\downarrow$ \\ 
\midrule
A & $\times$     & $\times$     & 22.0 & 43.4 & 53.5 & 8.0 \\
B & $\checkmark$ & $\times$     & 23.2 & 44.3 & 54.2 & 7.0 \\
C & $\times$     & $\checkmark$ & 23.6 & 45.5 & 54.4 & 7.0 \\
\rowcolor{ours-highlight}
D & $\checkmark$    & $\checkmark$ & \textbf{26.6} & \textbf{48.1} & \textbf{57.2}  & \textbf{6.0}     \\ 
\bottomrule
\end{tabular}%
}
\caption{Ablation study on the two-stage semantics grounding approach. Zero-shot text-to-video retrieval results on MSR-VTT are reported. Only the Lexicon branch is trained in this experiment.}
\label{tab:two_stage_training_strategy_ablation}
\end{table}

\subsection{Two-stage Semantics Grounding Ablation}
In this section, we solely train an individual lexicon branch to ablate the proposed two-stage semantics grounding (TSG) approach.

\vspace{0.4em}
\noindent\textbf{Overall TSG.}
Figure \ref{fig:Two_stage_semantics_grounding_ablation} shows the top-10 activated dimensions of different variants. In the baseline model (a) without TSG, the video falls into semantically irrelevant dimensions, and the grounding ability of the text encoder is mostly lost. Essentially, in the baseline model, lexicon representations degrade into high-dimensional latent representations. 
In contrast, semantically relevant dimensions are activated in model (d) which is trained using TSG.
Moreover, line A and D in Table \ref{tab:two_stage_training_strategy_ablation} show that employing TSG significantly improves the R@1 performance by 4.6\%, validating the effectiveness of the proposed TSG approach.

\vspace{0.4em}
\noindent\textbf{The Freezing Strategy.}
Comparing (a) and (b) in Figure \ref{fig:Two_stage_semantics_grounding_ablation}, we observe that freezing the text encoder in the first stage reduces the activation of semantically irrelevant dimensions. This demonstrates the freezing strategy prevents textual lexicon distributions from being corrupted by cross-modal alignment.
Line A and B in Table \ref{tab:two_stage_training_strategy_ablation} show that the freezing strategy improves retrieval performance.

\vspace{0.4em}
\noindent\textbf{The MLM Task.}
In (c) of Figure \ref{fig:Two_stage_semantics_grounding_ablation}, although the text encoder is frozen in the first stage, the absence of the MLM task in the second stage results in the loss of semantic constraints for the text, leading to the activation of some semantically irrelevant dimensions.
Comparing (c) and (d), we observe that the MLM task effectively suppresses the activation of semantically irrelevant dimensions, showing that MLM helps preserve textual semantics. Line C and D in Table \ref{tab:two_stage_training_strategy_ablation} 
show combing MLM with the freezing strategy largely improves the model's performance.

\begin{figure}
\centering
\includegraphics[width=0.95\linewidth]{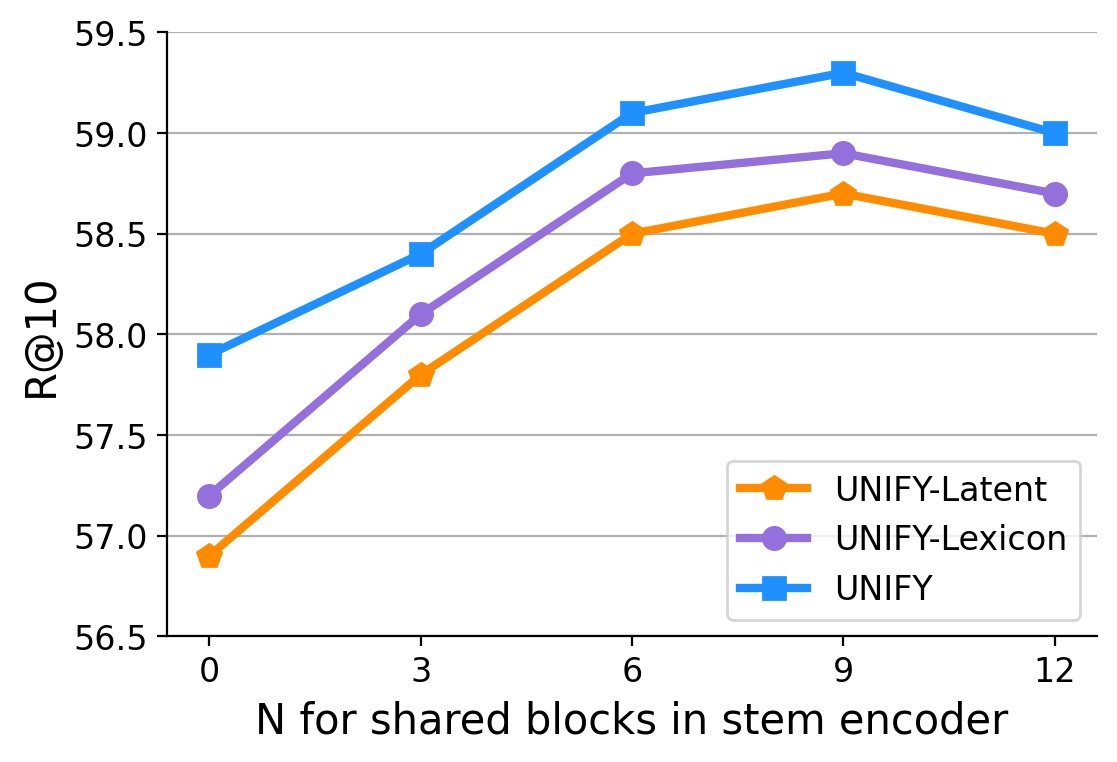}
\vspace{-8pt}
\caption{Effect of the number $N$ of shared blocks in the stem encoder. Zero-shot text-to-video retrieval results on MSR-VTT are reported. Here the experiments are conducted without self-distillation.}
\label{fig:N_stem_encoder}
\end{figure}

\subsection{Unified Learning Scheme Ablation}

\vspace{0.4em}
\noindent\textbf{Structure sharing.} Comparing line A and line B in Table~\ref{tab:unify_ablation}, we can find that sharing the stem encoder between the latent and lexicon branches leads to an improvement in performance for both branches, lifting 1.8\% and 1.7\% respectively. This highlights that sharing shallow layers between the two representation types can facilitate knowledge transfer and improve the representation ability of both branches. 
Moreover, performance boost is also observed in the fused UNIFY results, lifting from 26.8\% to 28.2\% in R@1.

\vspace{0.4em}
\noindent\textbf{The number of shared stem blocks.}
As shown in Figure~\ref{fig:N_stem_encoder}, as we gradually increase the number of shared blocks, the performance of both the latent and lexicon branches has improved.
The results demonstrate that they benefit from knowledge transfer via structure sharing.
And the optimal performance is obtained at $N=9$.
However, when $N$ continues to increase to 12, a performance decline occurs. This is because at this time the two branches completely share the encoder, and only the projection heads are different. Thus the model cannot fully learn the two specific representations. This result further verifies the rationality of the proposed structure sharing strategy.

\vspace{0.4em}
\noindent\textbf{Self-distillation.} When additionally introducing the self-distillation strategy from Line B to Line C, we observe further improvement in the performance of both branches and the overall UNIFY results.
Latent representations provide extra supervision information for lexicon representations, and the enhanced lexicon representations in turn inject fine-grained semantic knowledge into the stem encoder.
The experimental results demonstrate the proposed self-distillation strategy can facilitate mutual learning between the two representations types and improves the performance of both UNIFY-Latent and UNIFY-Lexicon.

\section{Conclusion}
In this work, we presented a novel UNIFY framework, which learns lexicon representations for fine-grained semantics capturing, and unifies latent and lexicon representations for cross-modal retrieval.
We proposed a two-stage semantics grounding approach to enable lexicon representations to reflect fine-grained semantic concepts.
As latent and lexicon representations have different focuses, we further proposed a unified learning scheme to leverage this complementarity.
Our method largely outperforms existing video-text retrieval methods, validating the effectiveness of lexicon representations and the unified learning scheme.

\section{Acknowledgements}
This work is supported by the National Key Research and Development Program of China (Grant No. 2022ZD0118501), Beijing Natural Science Foundation (Grant No. JQ21017, L223003, 4224093), the Natural Science Foundation of China (Grant No. 61972397, 62036011, 62192782, U2033210, 62202470), The Project of Beijing Science and technology Committee (Project No. Z231100005923046).

\section{Bibliographical References}\label{sec:reference}

\bibliographystyle{lrec-coling2024-natbib}
\bibliography{lrec-coling2024-example}


\end{document}